%% file: main.tex
\documentclass[10pt,twocolumn,letterpaper]{article}

\usepackage{cvpr}              

\input{preamble}

\definecolor{cvprblue}{rgb}{0.21,0.49,0.74}
\usepackage[pagebackref,breaklinks,colorlinks,allcolors=cvprblue]{hyperref}

\def\paperID{9048} 
\def\confName{CVPR}
\def\confYear{2026}

\title{Towards Pixel-Wise Anomaly Location for High-Resolution PCBA \\ via Self-Supervised Image Reconstruction}

\author{
Wuyi Liu$^{1,2,*}$, 
Le Jin$^{2,3,*}$,
Junxian Yang$^{2}$,
Yuanchao Yu$^{2}$,\\
Zishuo Peng$^{4}$,
Jinfeng Xu$^{1}$,
Xianzhi Li$^{1,\dagger}$, 
Jun Zhou$^{3,\dagger}$\\[0.4ex]
$^{1}$ Huazhong University of Science and Technology,
$^{2}$ Siemens AG,\\
$^{3}$ University of Electronic Science and Technology of China,
$^{4}$ Peking University\\[0.6ex]
{\tt\small
m202574039@hust.edu.cn,
le.jin@siemens.com,
yang\_junxian@mail.ustc.edu.cn
}\\
{\tt\small
yuanchao.yu@siemens.com,
2300010761@stu.pku.edu.cn,
jinfengxu.edu@gmail.com,
}\\
{\tt\small
xzli@hust.edu.cn,
jun.zhou.sg@ieee.org}
}

\begin{document}
\maketitle
\begin{abstract}
 Automated defect inspection of assembled Printed Circuit Board Assemblies (PCBA) is quite challenging due to the insufficient labeled data, micro-defects with just a few pixels in visually-complex and high-resolution images. 
 To address these challenges, we present \textbf{HiSIR-Net}, a \textbf{Hi}gh resolution, \textbf{S}elf-superv\textbf{I}sed \textbf{R}econstruction framework for pixel-wise PCBA localization. 
 Our design combines two lightweight modules that make this practical on real 4K-resolution boards: (i) a Selective Input-Reconstruction Gate (\textbf{SIR-Gate}) that lets the model decide where to trust reconstruction versus the original input, thereby reducing irrelevant reconstruction artifacts and false alarms; and (ii) a Region-level Optimized Patch Selection (\textbf{ROPS}) scheme with positional cues to select overlapping patch reconstructions coherently across arbitrary resolutions. Organically integrating these mechanisms yields clean, high-resolution anomaly maps with low false positive (FP) rate.
 To bridge the gap in high-resolution PCBA datasets, we further contribute a self-collected dataset named \textbf{SIPCBA-500} of 500 images.
We conduct extensive experiments on our SIPCBA-500 as well as public benchmarks,
demonstrating the superior localization performance of our method while running at practical speed.
Full code and dataset will be made available upon acceptance.

\vspace{-2mm}
\end{abstract}
\begin{figure}[t]
  \centering{
\includegraphics[width =\linewidth]{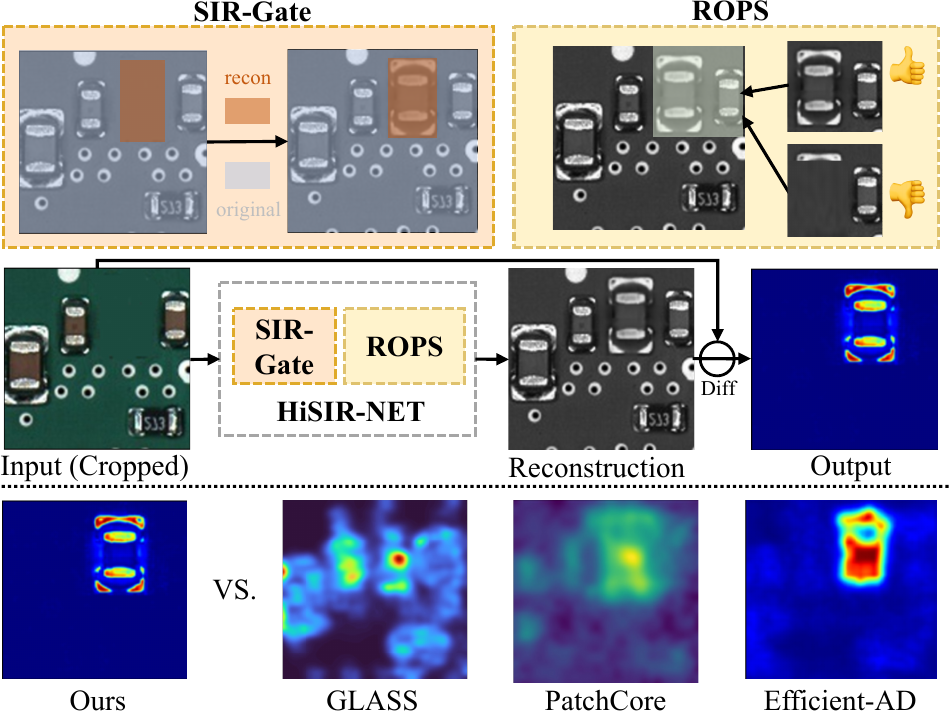}
}
\vspace{-3mm}
\captionof{figure}{
Our \textbf{HiSIR-Net} mainly consists of two novel modules: (i) SIR-Gate to suppress irrelevant reconstruction noise, and (ii) ROPS to select the best reconstruction from overlapping patches.
Clearly, our method significantly outperforms existing methods (e.g., GLASS\cite{10.1007/978-3-031-72855-6_3}, PatchCore\cite{roth2022towards} \& Efficient-AD\cite{batznerEfficientadAccurateVisual2024}).
}
\label{fig:intro}
\vspace{-3mm}
\end{figure}


\begingroup
\renewcommand\thefootnote{} 
\footnotetext{$^{*}$ Equal contribution.}
\footnotetext{$^{\dagger}$ Corresponding author.}
\footnotetext{This work was completed during Wuyi Liu's internship at Siemens AG.}
\endgroup
\section{Introduction}


Printed Circuit Board Assemblies (PCBA) serve as core components in consumer electronics, medical equipment, and beyond, with their quality being a decisive factor in end-product reliability.
Therefore, automated image-based defect inspection is pivotal in modern manufacturing. 
Compared with other industrial images, PCBA images present a regular circuit board layout (highly structured) at the global level, while at the local level, they feature dense arrangements of diverse components such as resistors, capacitors, and chips, along with fine reflective structures like solder joints and pins.
Hence, the precise localization of subtle anomalies on PCBA faces the following unique challenges.

First, \emph{data imbalance}. The severe disparity between the number of qualified and defective products renders standard supervised methods impractical, necessitating unsupervised/self-supervised approaches.
Second, \emph{micro-defects}. Local anomalies (e.g., missing components or pin deformation) are often just a few pixels in size, thus causing state-of-the-art methods like PatchCore~\cite{roth2022towards} and EfficientAD~\cite{batznerEfficientadAccurateVisual2024} to exhibit high false-positive rates; see Fig.~\ref{fig:intro} as examples. 
%
Third, \emph{high-resolution}. To clearly capture these tiny structures and anomalies, PCBA images are typically high-resolution up to 4K.
Unfortunately, naively applying state-of-the-art anomaly detection methods to 4K imagery either incurs prohibitive GPU memory and computation—resulting in impractically long training/inference—or loses sensitivity to few-pixel defects due to aggressive downsampling and coarse receptive fields.
Therefore, it is imperative to develop a scalable, unsupervised (or self-supervised), high-resolution anomaly detection framework offering both high precision and computational efficiency.

To address these challenges, we introduce \textbf{HiSIR-Net}, a \textbf{Hi}gh-resolution \textbf{S}elf-superv\textbf{I}sed \textbf{R}econstruction network designed for pixel-wise PCBA anomaly localization. Our framework incorporates two key innovations to overcome the limitations of prior work (see the top of Fig.\ref{fig:intro}). First, we propose an \textbf{Selective Input-Reconstruction Gate (SIR-Gate)} module. SIR-Gate acts as a spatial bottleneck that decides where to pass the original input versus the reconstruction. By filtering out reconstruction noise in benign regions, SIR-Gate substantially reduces false positives. Second, we introduce a \textbf{Region-level Optimized Patch Selection (ROPS)} scheme, enhanced with positional cues, to selectively merge overlapping patch reconstructions into a full-resolution output. This mechanism eliminates boundary artifacts and yields the consistency of reconstruction.


To facilitate rigorous evaluation and further research, we also present \textbf{SIPCBA-500}, a new 4K-resolution industrial benchmark dataset created specifically for this challenging task. It features a large training set of only normal samples and a meticulously annotated test set with pixel-level ground truth, directly addressing the gap left by existing datasets. On SIPCBA-500, HiSIR-Net achieves state-of-the-art localization accuracy while maintaining practical inference speeds, demonstrating its real-world viability. The main contributions of this paper are threefold:
\begin{itemize}
    \item A novel self-supervised network, \textbf{HiSIR-Net}, integrates Selective Input-Reconstruction Gate (SIR-Gate) and Region-level Optimized Patch Selection (ROPS) to achieve highly accurate, pixel-level anomaly localization on high-resolution PCBAs with a low false-positive rate.
    \item The introduction of \textbf{SIPCBA-500}, the first high-quality industrial benchmark for self-supervised PCBA inspection, featuring 4K resolution images, a normal-only training split, and pixel-precise annotations for realistic defect categories.
    \item Extensive experiments demonstrating that HiSIR-Net significantly outperforms state-of-the-art methods in both pixel-wise metrics and defect-level evaluation, validating its effectiveness for industrial deployment.
\end{itemize}

\section{Related Work}

\subsection{Anomaly Detection}

Machine learning methods for anomaly detection can be broadly categorized as reconstruction-based or embedding-based. Reconstruction-based methods typically employ autoencoders or generative models trained on normal data, using reconstruction errors to detect anomalies~\cite{batznerEfficientadAccurateVisual2024,Chen2024LGFDR,ZAVRTANIK2021107706}.  Embedding-based approaches rely on deep neural networks to cluster normal samples in feature spaces, thus isolating anomalies at inference time. Notable examples include PaDiM~\cite{defardPaDiMPatchDistribution2021}, Deep SVDD~\cite{yiPatchSvddPatchlevel2020}, PatchCore, and Student-Teacher frameworks like AST~\cite{rudolphAsymmetricStudentteacherNetworks2023} or EfficientAD~\cite{batznerEfficientadAccurateVisual2024}. Despite high performance, directly applying these methods to PCBA inspection poses unique challenges we address.

\subsection{Development in PCB \& PCBA Inspection}

PCB defects involve primarily 2D issues in bare board manufacturing—open circuits, extra copper, etc. Recent developments using supervised deep learning detection frameworks (e.g., YOLO series~\cite{xinPCBElectronicComponent2021, duYOLOMBBiPCBSurface2023, yangEnhancedDetectionMethod2023}, SSD~\cite{jiangMultitargetDetectionPcb2022}, and Transformer-based models~\cite{chenPCBDefectDetection2022, anLPViTTransformerBased2022}) achieved significant accuracy improvements. However, supervised methods demand large annotated datasets, hindered by imbalance issues, limiting their flexibility~\cite{xinPCBElectronicComponent2021}. To address these issues, unsupervised PCB defect detection methods have emerged, including recently proposed models such as SLLIP~\cite{YAO2023113611}, ReconPatch\cite{hyun2024reconpatch} and PCBSSD~\cite{LI2025116342}. However, inspection of PCBAs after component assembly poses additional complexity due to 3D objects, diverse defect types, variable reflections, and normal positional variance. Recent approaches using semantic segmentation~\cite{liIndustryorientedDetectionMethod2024} or thermal imaging~\cite{jeonContactlessPCBADefect2022} still struggle with inherent dataset imbalance and generalization challenges. Unsupervised/self-supervised methods become increasingly appealing~\cite{yaoPCBDefectDetection2023}, yet general-purpose anomaly detectors perform unsatisfactorily on PCBA inspection due to unique domain-specific aspects—ultra-high resolution, strong reflections, positional variance, which make unsupervised reconstruction and template-based approaches highly sensitive and prone to false positives. 




\subsection{Public Datasets}
Public benchmarks have driven progress in visual anomaly detection, with MVTec AD widely adopted for general industrial surfaces~\cite{defardPaDiMPatchDistribution2021, yiPatchSvddPatchlevel2020}. For circuit-board inspection specifically, commonly used datasets such as DeepPCB~\cite{tangOnlinePCBDefect2019} and HRIPCB~\cite{huangPCBDatasetDefects2019, huangHRIPCBChallengingDataset2020a} primarily target bare-board defects and rely on supervised settings or paired templates. These datasets typically provide low–to–medium resolution imagery, limited defect taxonomies, and do not capture the optical and structural complexity of assembled PCBs—e.g., 3D components and illumination variations. As a result, existing resources are ill-suited for evaluating self-supervised, pixel-level anomaly localization on assembled boards under normal-only training constraints. To address these gaps, we introduce SIPCBA-500, a 4K-resolution benchmark designed for normal-only training with pixel-accurate annotations and realistic defect distributions. Full details and statistics are provided in the  section~\ref{sec:dataset}.

\section{Method}
\label{ssnet}
 
\begin{figure*}[t]
  \centering{
    \includegraphics[width=\textwidth]{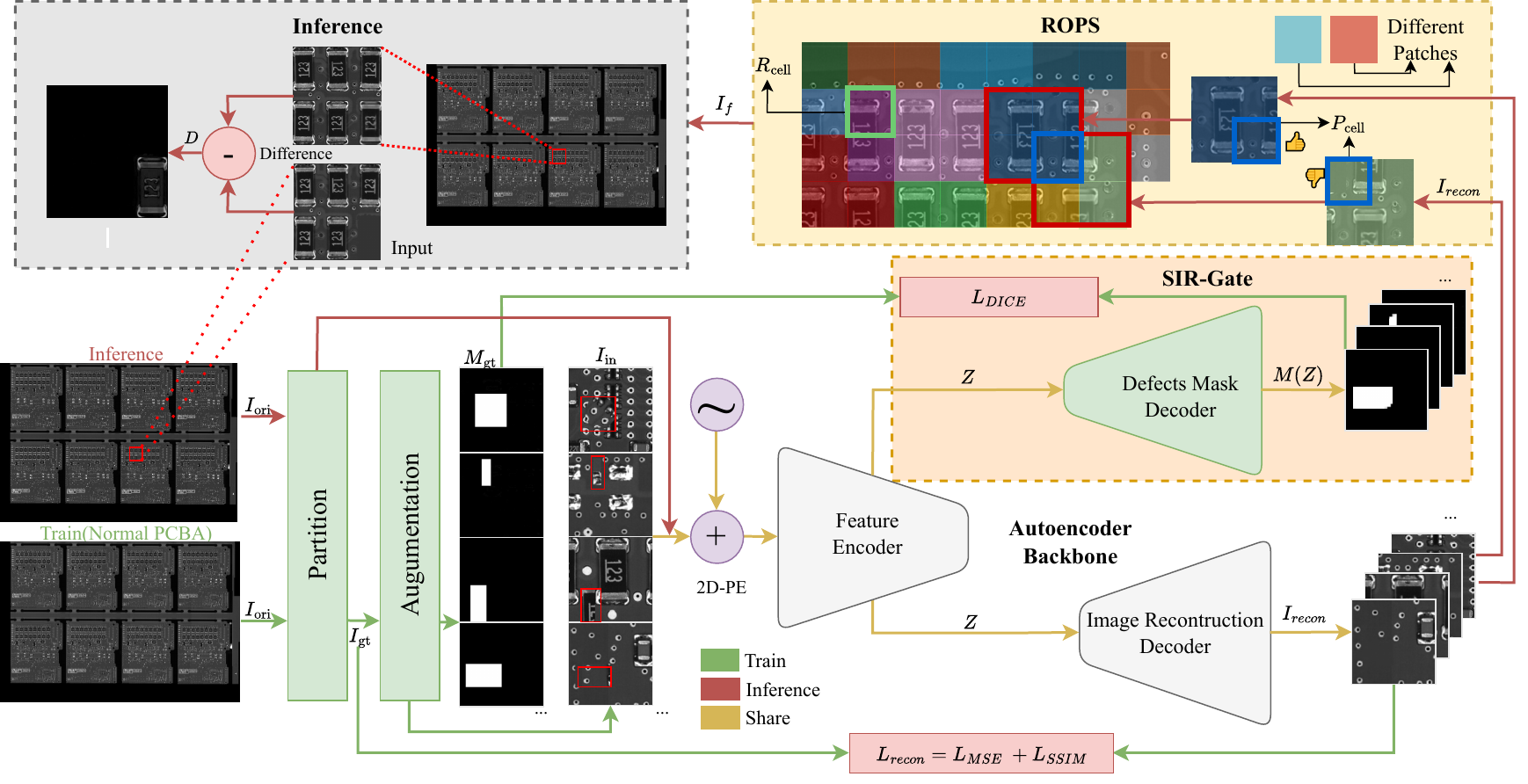}
  }
  \vspace{-3mm}
  \caption{Overall pipeline of HiSIR-Net with SwinUnet as backbone. In part ROPS, regions with the same color on PCBA are from the same patch, regions with different colors are from different patches.}
  \label{fig:pipeline}
\end{figure*}

HiSIR-Net explicitly addresses key challenges in pixel-level anomaly localization for industrial PCBA inspection, including extreme class imbalance (only normal samples available), high-resolution complexity and subtle variations. Specifically, as illustrated in Fig.~\ref{fig:pipeline}, our framework leverages three core innovations: (1) a self-supervised reconstruction framework specifically designed to address severe data imbalance; (2) a Selective Input-Reconstruction Gate (SIR-Gate) strategy that selectively reconstructs anomaly-relevant regions, significantly reducing irrelevant reconstruction noise; and (3) a Region-level Optimized Patch Selection (ROPS) enhanced by two-dimensional positional encoding, which resolves patch boundary discontinuities and spatial confusion among visually similar PCBA components. Integrating these complementary modules, HiSIR-Net robustly fulfills the stringent precision and recall demands of real-world PCBA inspection tasks.
\subsection{Self-Supervised Reconstruction Networks}
In response to the extreme data imbalance in industrial PCBA environments where nearly all available images are defect-free, we introduce a self-supervised anomaly detection framework that harnesses reconstruction as its linchpin.
By training exclusively on normal samples, our approach learns to model the normal appearance manifold with high fidelity, enabling abnormal regions to be identified through reconstruction discrepancies during inference.

Specifically, as shown in the bottom-left of  Fig.~\ref{fig:pipeline}, we first partition an ultra-high-resolution (i.e., 4K) PCBA image ($I_{ori}$) into $128$$\times$$128$ patches ($I_{gt}$) following standard practice to reduce computational complexity and facilitate localized feature extraction.
Next, to encourage the model to learn robust contextual understanding and meaningful representations without labeled data, we apply industrial-standard augmentation techniques to each input patch to mimic defects, including random masking, geometric transformations, and selective pixel drops ($I_{in}$).
Then we force the model to reconstruct normal inputs from simulated defective inputs, thereby enabling it to learn the distribution of normal samples.

The reconstruction objective combines mean squared error loss $\mathcal{L}_{MSE}$ and structural similarity loss $\mathcal{L}_{SSIM}$ in a hybrid formulation:
\begin{equation}
\label{eq:recon_loss}
\mathcal{L}_{recon} = \mathcal{L}_{MSE}(I_{recon}, I_{gt}) + \lambda \cdot \mathcal{L}_{SSIM}(I_{recon}, I_{gt}),
\end{equation}
where $I_{recon}$ and $I_{gt}$ are reconstructed and ground-truth images, respectively. The parameter $\lambda$ balances pixel-level accuracy and structural similarity; specifically, a larger $\lambda$ places greater emphasis on structural consistency, improving robustness to minor pixel-level variations, while a smaller $\lambda$ prioritizes strict pixel-wise reconstruction fidelity.

During inference, as shown in the top-left of Fig.~\ref{fig:pipeline}, pixel-level defect localization is achieved by computing the difference between the input PCBA $I_{ori}$ and its corresponding defect-free reconstruction $I_{f}$.
\begin{equation}
D = \left| I_{f} - I_{ori} \right|,
\label{eq:abs_diff_map}
\end{equation}
where $|\cdot|$ is applied element-wise, $I_{f}$ is the merged output produced by ROPS which will be introduced in section~\ref{rops}.

Unlike teacher-student paradigms (e.g., Efficient-AD~\cite{batznerEfficientadAccurateVisual2024}) heavily relying on pretrained external teacher models, or embedding-based methods (e.g., SLLIP~\cite{yaoPCBDefectDetection2023}) that depend on global feature-space consistency constraints, our reconstruction-based strategy explicitly models anomalies as pixel-wise deviations within the original image space. This formulation inherently ensures precise pixel-level localization and avoids domain mismatch introduced by external pretrained networks, while effectively addressing difficulties embedding-based methods encounter when detecting subtle visual anomalies.

\subsection{Selective Input-Reconstruction Gate}
\begin{figure}[t]
\centering{
  \includegraphics[width = \linewidth]{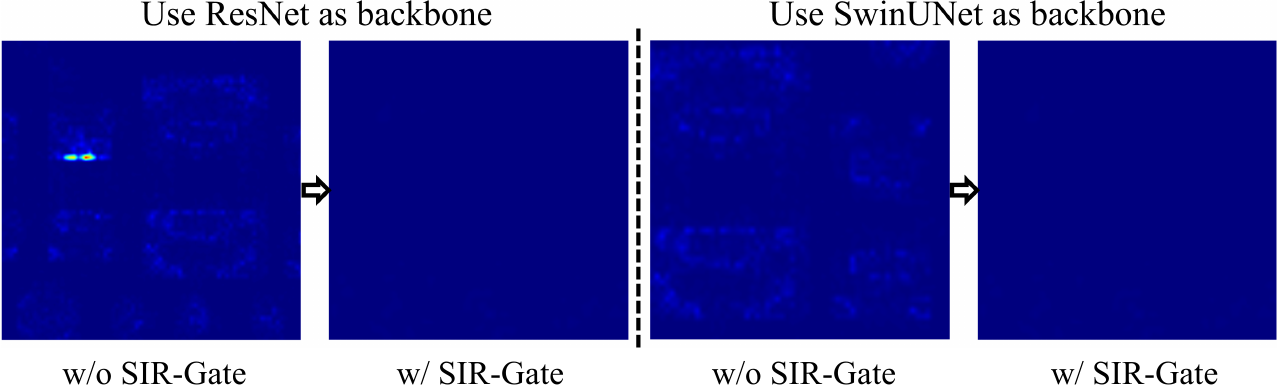}
  }
  \vspace{-4mm}
  \captionof{figure}{With the proposed SIR-Gate, reconstruction noise is greatly suppressed across different backbones.}
  \label{fig:mask_comparison}
\end{figure}
While reconstruction-based frameworks can effectively model normal appearance distributions, the reconstruction noise caused by irrelevant background details or minor normal variations often leads to false positives, especially on backbones such as ResNet~\cite{7780459}, as shown in Fig.~\ref{fig:mask_comparison}. 

To explicitly address this challenge, we propose \textbf{SIR-Gate} (Selective Input–Reconstruction Gate), a lightweight spatial gating mechanism that decides, per pixel, whether to trust the reconstruction or the original input. Concretely, as shown in Fig.~\ref{fig:pipeline}, after each augmented patch passes through the feature encoder, the latent representation $Z$ splits into two branches: one connected to the image reconstruction decoder and the other to a mask decoder to predict a spatial gate $M(Z)$. Where $M(Z)$ has near-zero values (likely normal regions), we discard the corresponding positions in the reconstructed image $I_{recon}$ and preserve those from the input $I_{in}$ to form the final output, thereby reducing irrelevant reconstruction artifacts. Conversely, for non-zero entries in $M(Z)$ (likely defect locations), we retain $I_{recon}$ to encourage the network to reconstruct normal patterns within suspected anomalous regions. Formally,
\begin{equation}
I_{recon} = M(Z) \odot I_{recon} + (1 - M(Z)) \odot I_{in}.
\label{eq:blend}
\end{equation}
This spatial fusion enforces the model to decide \emph{where} to reconstruct and \emph{where} to preserve, implicitly guiding it to focus on anomaly-relevant regions without introducing additional assumptions.

Different backbone architectures exhibit distinct reconstruction behaviors, which call for different relative strengths of reconstruction-driven versus mask-driven supervision. We unify training under a single composite objective and adapt across backbones by tuning the Dice loss weight:
\begin{equation}
\mathcal{L}_{total} = \mathcal{L}_{recon}(I_{recon}, I_{gt}) + \gamma \, \mathcal{L}_{dice}(M(Z), M_{gt}),
\label{eq:total}
\end{equation}
where $\mathcal{L}_{recon}$ is the hybrid MSE–SSIM loss defined in Eq.~\eqref{eq:recon_loss}, and 
\begin{equation}
    \mathcal{L}_{dice}(M(Z), M_{gt}) =1- \frac{2 \sum M(Z) M_{gt} + \epsilon}{\sum M(Z) + \sum M_{gt} + \epsilon},
\end{equation}
regularizes the mask prediction branch using the pseudo ground-truth mask $M_{gt}$ produced by the augmentation pipeline in Section~\ref{ssnet}.

For convolutional backbones such as ResNet, reconstructions tend to exhibit noticeable pixel-level noise, and the fusion gradient in Eq.~\eqref{eq:blend} provides a strong supervisory signal to learn accurate masks. In this regime, we set $\gamma$ close to zero (effectively allowing fusion-driven supervision to dominate), so $M(Z)$ is shaped primarily by reconstruction fidelity while Dice serves as a negligible regularizer.

In contrast, transformer-based backbones such as SwinUNet produce smoother, low-noise reconstructions, making fusion-derived gradients weaker for mask learning. Here, we increase $\gamma=0.1$ to strengthen explicit mask supervision and ensure the gate remains responsive to subtle anomaly cues. In practice, moderate $\gamma$ values provide a good balance: fusion retains its role in driving reconstruction fidelity, while Dice complements it by stabilizing mask learning without overwhelming the reconstruction signal.

In summary, SIR-Gate uses a single training objective (Eq.~\eqref{eq:total}) and a single inference rule (Eq.~\eqref{eq:blend}), adapting to different backbones by adjusting the Dice weight $\gamma$: small $\gamma$ for high-noise convolutional backbones where fusion is sufficiently informative, and larger $\gamma$ for low-noise transformer backbones where explicit mask regularization is beneficial. As visualized in Fig.~\ref{fig:mask_comparison}, this weight-controlled adaptation consistently suppresses irrelevant reconstruction noise and enhances anomaly selectivity across architectures.

\subsection{Region-level Optimized Patch Selection}
\label{rops}
\begin{figure}[t]
\centering{
  \includegraphics[width=\linewidth]{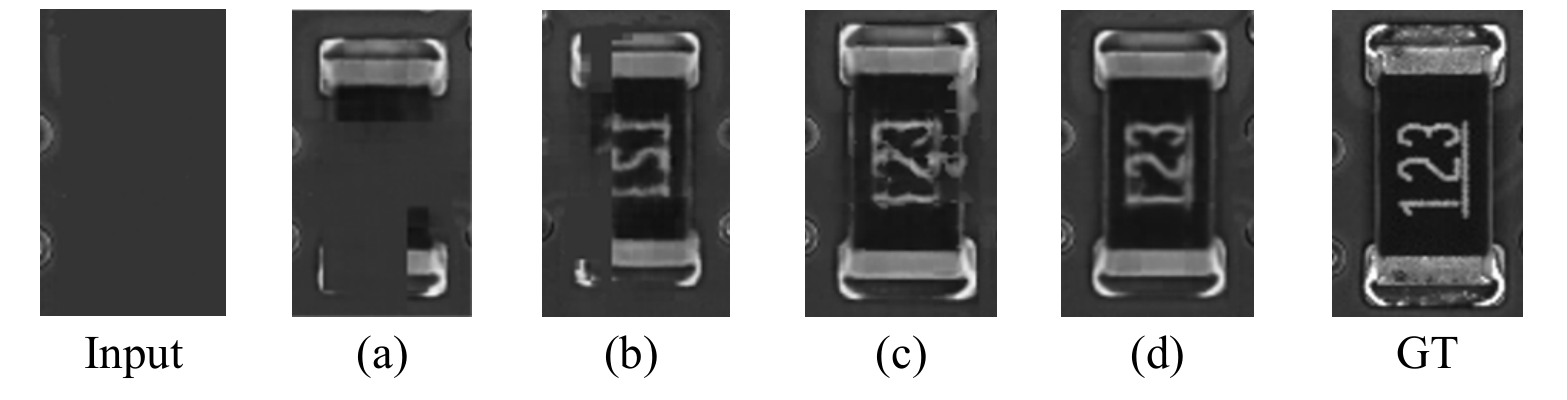}
  }
  \vspace{-5mm}
  \captionof{figure}{Visual comparison of reconstruction results under different patch combination strategies: (a) without positional encoding or overlaps, (b) with positional encoding only, (c) with overlap and pixel-level merging, (d) with overlap and ROPS.}
  \label{fig:patch_selection_comparison}
\end{figure}

\begin{algorithm}[t]
\caption{Region-level Optimized Patch Selection}
\label{algo:adaptive_selection}
\begin{algorithmic}[1]
\Require Original input patch $I_{in}$; set of reconstructed patches $\{P_k\}$ from $I_{recon}$; difference threshold $T_{\text{diff}}$.
\Ensure Final reconstructed image $I_{\text{f}}$.
\State Initialize $I_{\text{f}}$ as an empty canvas.
\State Partition each patch into four equal-sized quarter-patch regions.
\For{each region $R_{\text{cell}}$ in the image grid}
    \State Obtain the set $\mathcal{P}_{\text{cell}}$ containing all overlapping reconstructed patches for $R_{\text{cell}}$.
    \State Compute reconstruction difference:
    \[
        D_{\text{max}} = \max_{P_i \in \mathcal{P}_{\text{cell}}} 
        \left| I_{\text{in}}(R_{\text{cell}}) - I_{\text{recon}}^{P_i}(R_{\text{cell}}) \right|
    \]
    \If{$D_{\text{max}} < T_{\text{diff}}$}
        \State \Comment{Average overlapping regions}
        \State $I_{\text{f}}(R_{\text{cell}}) \gets 
        \text{Avg}_{P_i \in \mathcal{P}_{\text{cell}}}
        \{ I_{\text{recon}}^{P_i}(R_{\text{cell}}) \}$
    \Else
        \State \Comment{Potential anomaly region—choose maximum discrepancy patch}
        \State $P^* \gets \arg\max_{P_i \in \mathcal{P}_{\text{cell}}} 
        \left| I_{\text{in}}(R_{\text{cell}}) - I_{\text{recon}}^{P_i}(R_{\text{cell}}) \right|$
        \State $I_{\text{f}}(R_{\text{cell}}) \gets I_{\text{recon}}^{P^*}(R_{\text{cell}})$
    \EndIf
\EndFor
\State \Return $I_{\text{f}}$
\end{algorithmic}
\end{algorithm}

While SIR-Gate significantly enhances anomaly selectivity by suppressing irrelevant reconstruction noise, residual artifacts still emerge along patch boundaries—particularly in challenging scenarios involving missing or shifted components (Fig.~\ref{fig:patch_selection_comparison}(a)). This indicates that spatial ambiguities arise from visually similar components appearing at distinct positions across PCBA images. Explicit positional encoding is thus introduced as an initial attempt, where we embedded two-dimensional positional cues\cite{MaskedAutoencoders2021} directly into the network. Though this positional encoding approach considerably reduces spatial ambiguities by explicitly providing global positional context (Fig.~\ref{fig:patch_selection_comparison}(b)), it did not entirely resolve issues related to boundary inconsistencies or subtle discrepancy confusion.

We further propose an overlapping patch partitioning strategy, sampling patches at half the patch-size intervals to ensure comprehensive coverage of each region. However, standard patch-merging methods, such as pixel-level averaging (Fig.~\ref{fig:patch_selection_comparison}(c)), either introduce reconstruction noise or fail to maintain global consistency, thus amplifying FP.

To optimally balance precision and sensitivity, we introduce a novel Region-level Optimized Patch Selection (ROPS) approach. Specifically, each overlapping patch is logically divided into four equal-sized regions, resulting in each region being potentially covered by up to four patches (illustrated visually in the right-top of Fig.~\ref{fig:pipeline}). The core idea behind ROPS is to adaptively select regions based on reconstruction differences. If all overlapping patches within a region exhibit reconstruction differences (with respect to the original input) below a defined threshold ($T_{diff}=0.1$), we average all overlapping regions. Otherwise, if any overlapping patch exceeds the threshold, indicating a potential anomaly, we select the patch region with the greatest reconstruction difference—highlighting maximum anomaly cues. Formally, 
{\small
\begin{equation}
I_{f}(R_{cell}) = 
\begin{cases}
I_{recon}^{P^*}(R_{cell}), \text{if} \max\limits_{P_i \in \mathcal{P}_{cell}}\text{diff}(P_i,R_{cell}) \geq T_{diff} \\
\text{Avg}_{P_i \in \mathcal{P}_{cell}}\{I_{recon}^{P_i}(R_{cell})\},  \text{otherwise}
\end{cases}
\end{equation}
}
where
\begin{equation}
P^* = \arg\max\limits_{P_i \in \mathcal{P}_{cell}}\text{diff}(P_i,R_{cell})
\end{equation}
and $\text{diff}(\cdot)$ denotes the reconstruction difference between patch $P_i$ and the original image $I_{in}$ within region $R_{cell}$.

Algorithm~\ref{algo:adaptive_selection} clearly illustrates how ROPS effectively balances global coherence and sensitivity. As visually demonstrated in Fig.~\ref{fig:patch_selection_comparison}(d) and validated further in experiments, integrating ROPS effectively eliminates boundary artifacts and significantly improves precision.

\section{Experiments}

\begin{figure*}[t]
  \centering{
    \includegraphics[width=\textwidth]{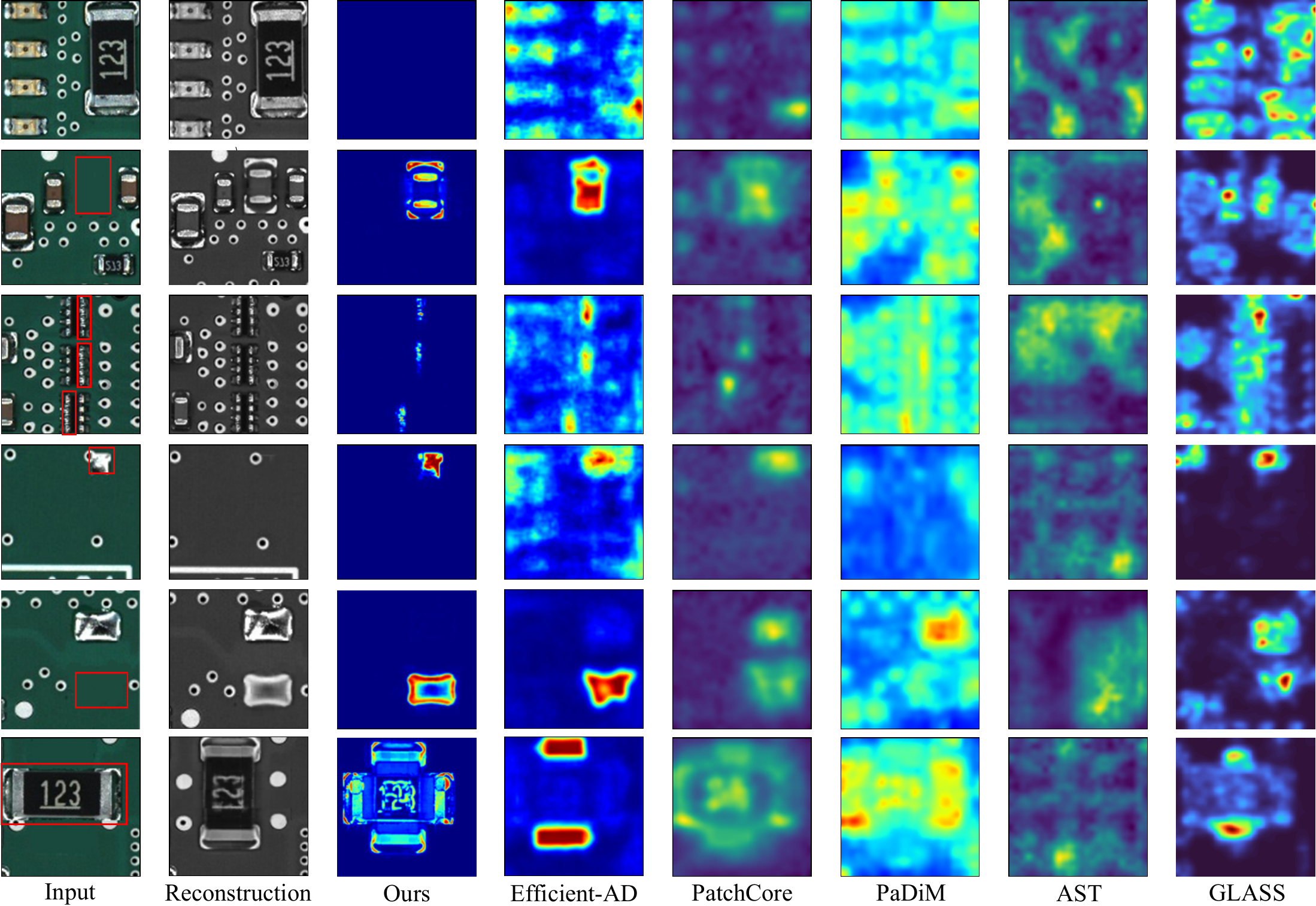}
  }
  \vspace{-3mm}
  \caption{Qualitative comparison results show that HiSIR-Net produces less background noise and clearer anomalies. Note that our model operates on the original full-resolution PCBA images; for visualization clarity, we display cropped sub-images from these large inputs.}
  \label{fig:Comparasion}
\end{figure*}

\subsection{Benchmarking Dataset: SIPCBA-500}\label{sec:dataset}
\begin{table}[t]
\centering
\footnotesize
\caption{
Defect category statistics in the SIPCBA-500 test split.
}
\vspace{-1mm}
\begin{tabular}{lcc}
\toprule
\textbf{Defect Type} & \textbf{Count} & \textbf{Percentage (\%)} \\
\midrule
Missing component              & 174 & 57.0 \\
Solder bridging (short)        & 85  & 27.9 \\
Foreign object debris (FOD)    & 34  & 11.1 \\
Extra component                & 9   & 3.0 \\
Misalignment                   & 3   & 1.0 \\
\midrule
\textbf{Total}                 & \textbf{305} & \textbf{100.0} \\
\bottomrule
\end{tabular}
\label{tab:dataset_stats}
\vspace{-3mm}
\end{table}
Existing industrial datasets like DeepPCB~\cite{tang2019online} and HRIPCB~\cite{huangHRIPCBChallengingDataset2020a} focus on bare-board defects and require supervision, overlooking the complexities of \textbf{assembled PCBs} where reflections, 3D components, and misalignments are common. This creates a critical gap, as there is no public benchmark for evaluating modern self-supervised methods on assembled boards with pixel-level ground truth.

To bridge this gap, we introduce \textbf{SIPCBA-500}, the first high-quality benchmark for self-supervised PCBA anomaly localization, defined by four core properties:
\begin{itemize}
    \item \textbf{Resolution and Realism:} Native 4K-resolution images from a production line, every image consists 8 PCBAs that preserve natural variations in lighting, specular highlights, and component pose.
    \item \textbf{Normal-only Training Set:} 450 defect-free images for learning the normal appearance manifold, aligning with real-world data constraints.
    \item \textbf{Pixel-level Annotated Test Set:} A test set of 50 images, each with a defect-free version and a paired defect-injected version. All 305 defects are annotated with pixel-precise polygons.
    \item \textbf{Realistic Defect Distribution:} The test set mirrors industrial defect frequencies across five common categories (Table~\ref{tab:dataset_stats}), including rare but critical faults.
\end{itemize}
By providing paired, pixel-accurate, and industrially validated data, SIPCBA-500 enables rigorous and fair evaluation of both localization precision and false-positive robustness. The dataset and its annotations will be publicly released upon acceptance to facilitate future research.

\subsection{Experimental Setup}
We compared HiSIR-Net's performance with several established SOTA baselines: ResNet\cite{7780459}, AST\cite{rudolphAsymmetricStudentteacherNetworks2023}, PaDiM\cite{defardPaDiMPatchDistribution2021}, RIAD\cite{zavrtanik2021reconstruction}, PatchCore\cite{roth2022towards}, EfficientAD\cite{batznerEfficientadAccurateVisual2024}, GLASS\cite{10.1007/978-3-031-72855-6_3}and PatchSVDD\cite{yiPatchSvddPatchlevel2020}. Most of these baselines do not natively support ultra-high-resolution data, thus we slightly modified their data-loaders to accommodate large-scale image patches; all other hyperparameters and training strategies strictly adhered to their respective official implementations. All experiments were conducted on a system equipped with a single NVIDIA RTX 3090 GPU and 32G RAM.

For HiSIR-Net training, we implement our experiment on backbones Resnet and SwinUNet, if not specially mentioned, HiSIR-Net refers to HiSIR-Net with SwinUNet as backbone. The batch size is set to 128 and random seed is set to 42 to ensure experimental reproducibility. Moreover, the hyperparameter $\lambda$ in our reconstruction loss \ref{eq:recon_loss} and $\gamma$ in total loss \ref{eq:total} is set to $0.1$ (for ResNet $\gamma=0.001$) based on validation performance.

During testing, predictions (generation of reconstructed images) were independently computed for each patch. Subsequently, all patch-level raw prediction results were spatially concatenated to reconstruct the complete PCBA-level output. Then we applied min-max normalization to the combined result to standardize anomaly scores for subsequent evaluations. These normalized PCBA-level predictions were then directly compared with the original ground-truth labels to compute pixel-wise AUROC and AUPRO metrics.

\subsection{Qualitative and Quantitative Results}
\begin{table}[t]
\centering
\footnotesize
\setlength{\tabcolsep}{4.5pt}
\renewcommand{\arraystretch}{1.15}
\caption{Quantitative comparison of pixel-wise AUROC, AUPRO, and inference time on the SIPCBA-500 dataset.}

\begin{tabular}{l c >{\columncolor{gray!15}}c c}
\toprule
\textbf{Method} & \textbf{AUROC} & \textbf{AUPRO} & \textbf{Time (s/img)} \\
\midrule
HiSIR-Net (ResNet) & 0.8932 & 0.7921 & 3.25 \\
AST (WACV2023) & 0.6956 & 0.6213 & 137.7 \\
PaDiM (ICPR2021) & 0.7628 & 0.6987 & 21.3 \\
PatchCore (CVPR2022) & 0.8463 & 0.7494 & 93.1 \\
EfficientAD (WACV2024) & 0.8611 & 0.7835 & 12.8 \\
PatchSVDD (ACCV2020) & 0.8296 & 0.6958 & 81.9 \\
GLASS (ECCV2024) & 0.8353 & 0.7332 & 38.7 \\
RIAD (Pattern Recognit.2021) & 0.8627 & 0.7838 & 24.1\\
\textbf{HiSIR-Net (SwinUNet)} & \textbf{0.9867} & \textbf{0.9512} & \textbf{2.78} \\
\bottomrule
\end{tabular}

\label{tab:results}
\end{table}
Table~\ref{tab:results} summarizes the quantitative analysis on SIPCBA-500. For industrial applications, minimizing false positives is critical, making the AUPRO metric—which penalizes poor precision at low recall rates—particularly relevant. HiSIR-Net achieves the highest pixel-wise AUROC (0.9867) and AUPRO (0.9512), while also delivering the fastest inference time (2.78 s/image), demonstrating its superior accuracy, reliability, and efficiency.

\begin{figure}[t]
  \centering{
  \includegraphics[width=\linewidth]{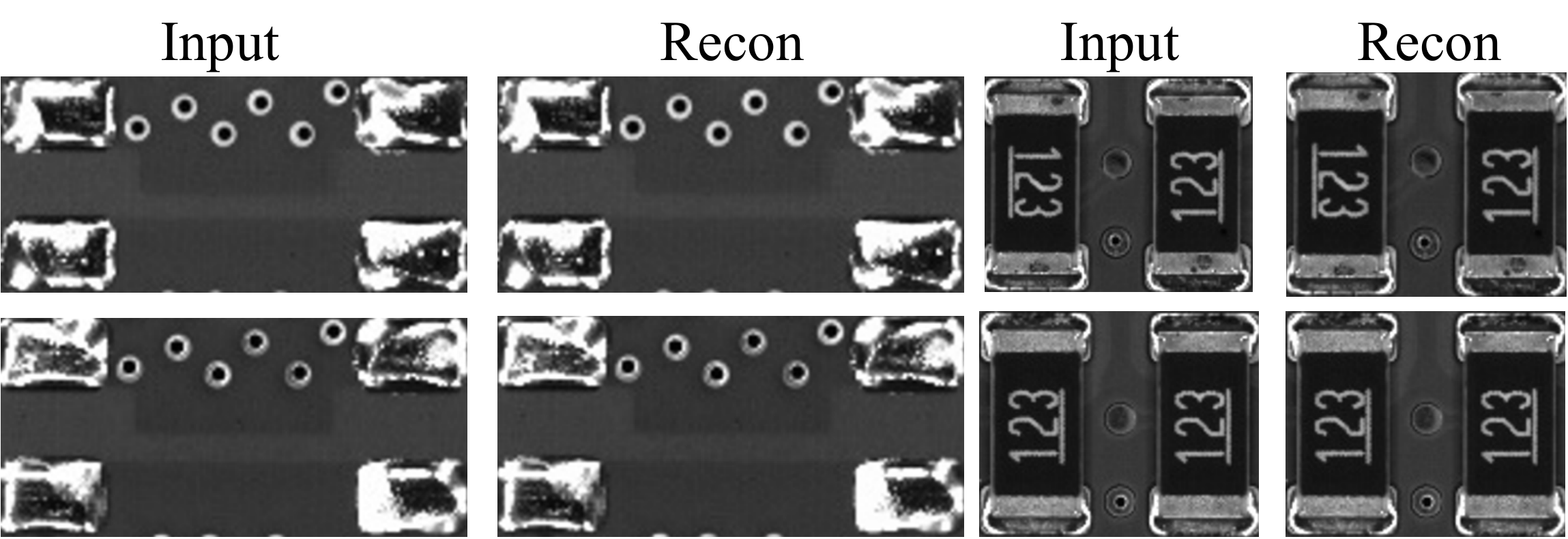}
  }
  \vspace{-3mm}
  \captionof{figure}{Reconstruction results under varying illumination and different rotation. Rows correspond to identical PCB locations captured under different lighting conditions.}
  \label{fig:light comparasion}
\end{figure}

Qualitatively, Fig.~\ref{fig:Comparasion} and Fig.~\ref{fig:light comparasion} provide a visual comparison. To ensure fairness, all heatmaps show raw anomaly scores after per-image min-max normalization, with no post-processing or thresholding applied. The visualizations transparently show that HiSIR-Net produces exceptionally clean backgrounds and sharply defined anomalies. This is in contrast to baselines like PatchCore and EfficientAD, which exhibit significant background noise due to their sensitivity to benign variations like illumination changes. This low-noise performance is a direct result of our proposed SIR-Gate and ROPS modules, not visualization artifacts. Fig.~\ref{fig:Comparasion} and Fig.~\ref{fig:light comparasion} consistently illustrate this property across different defect categories and lighting conditions.
\section{Analysis}
\label{sec:Analysis}
\subsection{Defect-Level Industrial Reliability Analysis}
To evaluate industrial readiness, we move beyond pixel-level AUROC to a stricter \textbf{defect-level evaluation} that prioritizes minimizing false positives (FPs). Standard anomaly detectors are typically optimized for pixel-wise metrics and produce dense heatmaps requiring extensive, dataset-specific tuning to yield binary decisions. As they are not designed for this low-FP, object-level task, we focus this analysis on HiSIR-Net's operational performance under a demanding industrial protocol.

Our evaluation follows three strict rules: 1) A single, global threshold is used to binarize all anomaly maps. 2) A defect is counted as a true positive (TP) only if its Intersection-over-Union (IoU) with a gt mask exceeds 0.5. 3) False positives are measured on an independent set of 50 defect-free images to simulate real-world conditions.

Under this protocol, HiSIR-Net (SwinUNet) achieves near-perfect robustness, generating only \textbf{one false positive} across the entire defect-free set. As shown in Table~\ref{tab:precision_recall}, it also attains a balanced defect-level precision and recall above 92\%, confirming its suitability for deployment. In contrast, the ResNet-based variant struggles with higher reconstruction noise, leading to lower performance.

\begin{table}[t]
\centering
\footnotesize
\captionof{table}{
\textbf{Defect-level precision and recall on SIPCBA-500 (IoU=0.5).} 
Existing SOTA methods are omitted since they target pixel-level AUROC rather than defect-level, low-FP evaluation, making direct comparison infeasible and misleading.
}
\renewcommand{\arraystretch}{1.05}
\setlength{\tabcolsep}{5pt}
\begin{tabular}{l c c}
\hline
Method & Precision (\%) & Recall (\%) \\
\hline
\rowcolor{gray!15} Efficient-AD & 35.63 & 57.58 \\
HiSIR-Net (ResNet) & 58.49 & 61.55 \\
HiSIR-Net (SwinUNet) & \textbf{93.26} & \textbf{92.37} \\
\hline
\end{tabular}
\label{tab:precision_recall}
\end{table}

\subsection{Generalization Analysis}

Although HiSIR-Net is explicitly designed for PCBA inspection, we additionally examine its cross-domain generalization using the widely adopted MVTec AD benchmark~\cite{bergmann2019mvtec}. 
It is important to note that the two domains differ fundamentally in both geometry and data priors. 
PCBA imagery exhibits a highly \textbf{planar and structured layout}: components are densely arranged on a flat surface, and positional variance across samples is minimal, typically limited to sub-pixel shifts caused by mechanical tolerances. 
In contrast, MVTec scenes contain \textbf{three-dimensional, unaligned objects} with significant viewpoint and pose variations. 
Such differences place MVTec largely outside the distribution for which HiSIR-Net’s reconstruction-based learning is optimized.

\begin{figure}[t]
  \centering{
  \includegraphics[width=\linewidth]{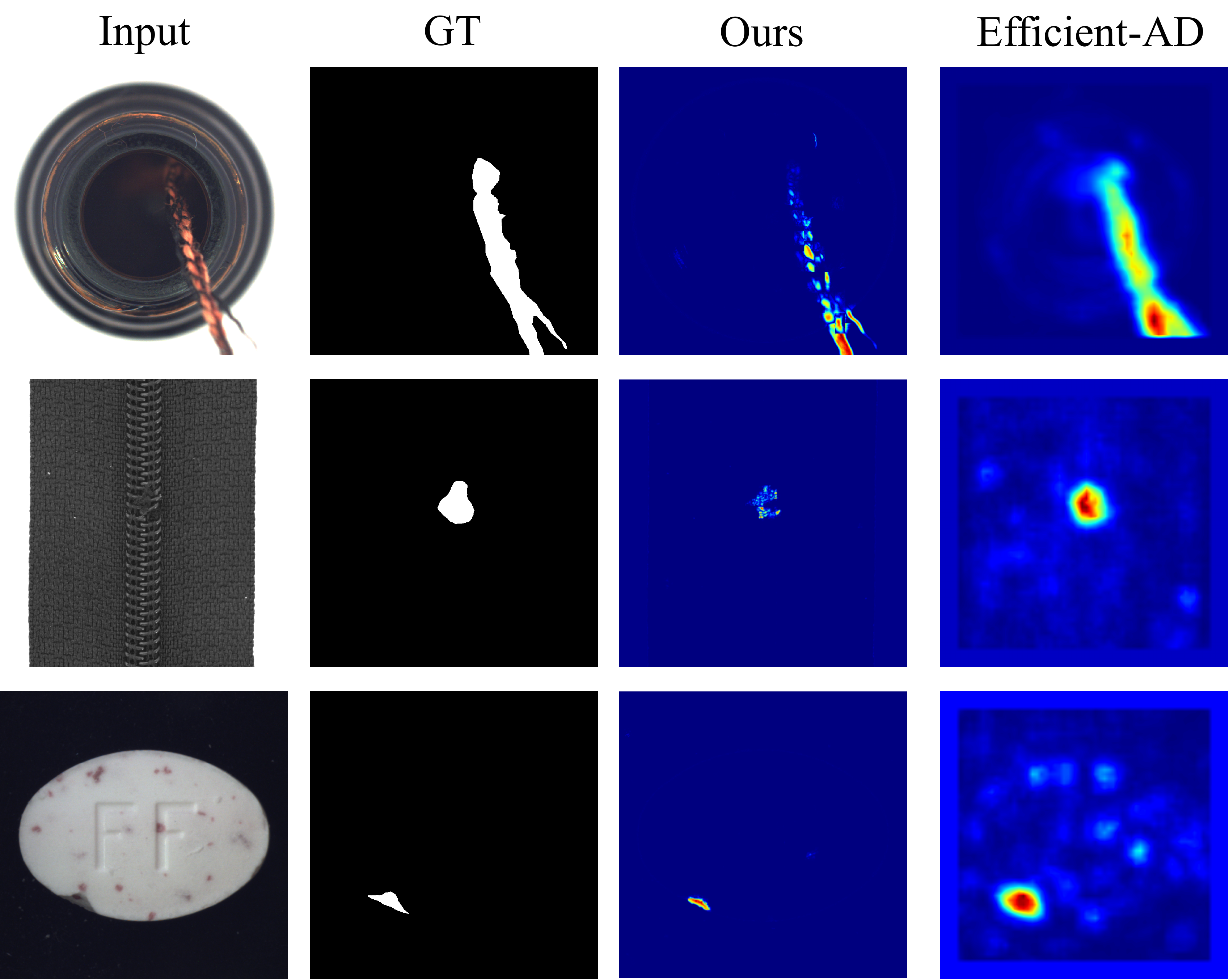}
  }
  \vspace{-3mm}
  \captionof{figure}{Inference results on MVTec dataset.}
  \label{fig:MVTec}
\end{figure}

Despite this severe domain gap, Fig.~\ref{fig:MVTec} shows that HiSIR-Net still produces stable, low-noise anomaly maps with clearly delineated defect regions.

This observation highlights a key insight: HiSIR-Net’s strength lies in modeling fine-grained, reflectance-sensitive features characteristic of PCBAs, but its inductive bias is less suited to volumetric, non-aligned scenes where global structure varies dramatically. 
Currently, there is no publicly available \textbf{unsupervised or self-supervised PCBA benchmark} other than SIPCBA-500, so MVTec remains the only external dataset for assessing cross-domain robustness. 
We expect that future research extending HiSIR-Net with domain-adaptive normalization or geometry-aware reconstruction priors could further enhance its generalization to out-of-plane industrial inspection tasks.

\subsection{Ablation Study}\label{sec:ablation}

To quantitatively validate the contribution of each module in HiSIR-Net, we conduct an ablation study focusing on three core components: (1) Region-level Optimized Patch Selection (\textbf{ROPS}), (2) two-dimensional positional encoding (\textbf{2D PE}), and (3) Selective Input–Reconstruction Gate (\textbf{SIR-Gate}). 
Table~\ref{tab:ablation_study} summarizes the results on the SIPCBA-500 dataset using both SwinUNet and ResNet backbones.

\begin{table}[t]
\centering
\footnotesize
\captionof{table}{
\textbf{Ablation study on SIPCBA-500.}
Precision/Recall (\%) under SwinUNet and ResNet backbones.
Symbols: \checkmark = module enabled; \(\times\) = module disabled; N/A = Without SIR-Gate, ResNet collapses into heavy reconstruction noise, making downstream defect localization infeasible.
}
\renewcommand{\arraystretch}{1.05}
\setlength{\tabcolsep}{3pt}
\begin{tabular}{c|ccc|cc|cc}
\hline
{No.} & \multicolumn{3}{c|}{Modules} &
\multicolumn{2}{c|}{SwinUNet} &
\multicolumn{2}{c}{ResNet} \\
\cline{2-8}
& ROPS & 2DPE & SIR-Gate & Precision & Recall & Precision & Recall \\
\hline
1 & \(\times\) & \(\times\) & \(\times\) & 81.2 & 84.4 & N/A & N/A \\
2 & \checkmark & \(\times\) & \(\times\) & 87.3 & 88.2 & N/A & N/A \\
3 & \checkmark & \checkmark & \(\times\) & 89.2 & 91.0 & N/A & N/A \\
4 & \checkmark & \(\times\) & \checkmark & 92.9 & 91.6 & 51.2 & 54.9 \\
5 & \(\times\) & \checkmark & \(\times\) & 82.9 & 85.5 & N/A & N/A \\
6 & \(\times\) & \(\times\) & \checkmark & 88.1 & 87.4 & 50.3 & 53.3 \\
7 & \(\times\) & \checkmark & \checkmark & 91.9 & 91.3 & 50.2 & 51.1 \\
8 & \checkmark & \checkmark & \checkmark & \textbf{93.3} & \textbf{92.4} & \textbf{58.5} & \textbf{61.6} \\
\hline
\end{tabular}
\label{tab:ablation_study}
\end{table}

\textbf{Effect of ROPS.}
Removing ROPS causes a clear precision drop ($-6.1\%$) due to boundary artifacts at patch seams. 
ROPS enforces spatial consensus through region-level selection, which merges overlapping patches coherently and suppresses local discontinuities in the reconstruction map.

\textbf{Effect of 2D Positional Encoding.}
2D positional encoding alone yields moderate improvement, but becomes much more effective when combined with ROPS (Row 3 vs. Row 5 in Table~\ref{tab:ablation_study}). 
This synergy arises because PE provides consistent spatial reference across patches, while ROPS leverages this alignment to enforce regional coherence.

\textbf{Effect of SIR-Gate.}
The impact of SIR-Gate varies significantly across backbone families.
For CNN-based backbones such as ResNet, removing SIR-Gate results in extremely noisy reconstruction where defect and background regions are visually indistinguishable, making defect-level analysis or threshold-based localization impossible. 
SIR-Gate stabilizes training by filtering irrelevant reconstruction signals through its mask-guided fusion, effectively turning an otherwise unusable model into a functional one.
In contrast, for Transformer-based SwinUNet, which inherently produces smoother and less noisy reconstructions, SIR-Gate primarily acts as a precision regularizer: it further suppresses minor false positives and sharpens defect boundaries. 
This behavior validates SIR-Gate’s backbone-adaptive property, serving as a necessary stabilizer for CNNs and a fine-tuning enhancer for Transformers.

\section{Conclusion}

We introduced HiSIR-Net, a high-resolution self-supervised reconstruction framework for pixel-wise PCBA defect localization under extreme label scarcity and few-pixel anomalies. The design pivots on two lightweight modules: (i) SIR-Gate, which selectively trusts input versus reconstruction to suppress false positives, and (ii) ROPS, a region-level selector that yields coherent 4K-resolution outputs from overlapping patches. We further release SIPCBA-500, a 4K-resolution dataset with normal-only training set and pixel-annotated testing set. Experiments on SIPCBA-500 and public benchmarks demonstrate our method can produce superior localization with low false positives rate at practical inference speed.

Despite our method's per-image runtime (2.78 s/4K image) achieves state-of-the-art inference speed among existing approaches and meets the lower bound of real-time usability, higher-throughput lines would still benefit from further model compression or hardware acceleration. Furthermore, like other PCB-specific methods, our framework is primarily optimized for PCBA characteristics, possibly limiting performance when generalized to significantly different inspection domains without additional adaptations.

\section*{Acknowledgment}
This work was supported by the National Key Research and Development Program for Young Scientists of China (2024YFB3310100), the Natural Science Foundation of Hubei Province (No. 2025AFB592), and the Natural Science Foundation of Wuhan (No. 2025040601020216). This work was also sponsored by Siemens AG.
{
    \small
    \bibliographystyle{ieeenat_fullname}
    \bibliography{main}
}
\end{document}

%% file: preamble.tex
\usepackage{algorithm}
\usepackage{algpseudocode}
\usepackage{booktabs}
\usepackage[table]{xcolor} 

